\documentclass{article}

\usepackage[nonatbib, preprint]{nips_2018}

\usepackage[utf8]{inputenc} 
\usepackage[T1]{fontenc}    
\usepackage{hyperref}       
\usepackage{url}            
\usepackage{booktabs}       
\usepackage{amsfonts}       
\usepackage{nicefrac}       
\usepackage{microtype}      
\usepackage{graphicx} 
\usepackage{color}

\usepackage[square,sort,comma,numbers]{natbib}
\title{Blindfold Baselines for Embodied QA}

\author{
    Ankesh Anand$^{1}$ \hspace{0.2pc}
    Eugene Belilovsky$^1$ \hspace{0.2pc}
    Kyle Kastner$^1$ \hspace{0.2pc}
    Hugo Larochelle$^{2,1}$ \hspace{0.2pc}
    Aaron Courville$^{1,3}$ \\[0.05in]
    {$^1$Mila} \hspace{0.5pc}
    {$^2$Google Brain}\hspace{0.5pc}
    {$^3$CIFAR Fellow} \\
}

\begin{document}

\maketitle

\begin{abstract}

We explore blindfold (question-only) baselines for Embodied Question Answering. The EmbodiedQA task requires an agent to answer a question by intelligently navigating in a simulated environment, gathering necessary visual information only through first-person vision before finally answering. Consequently, a blindfold baseline which ignores the environment and visual information is a degenerate solution, yet we show through our experiments on the EQAv1 dataset that a simple question-only baseline achieves state-of-the-art results on the EmbodiedQA task in all cases except when the agent is spawned extremely close to the object. 
\end{abstract}

\section{Introduction}
Recent breakthroughs in static, unimodal tasks such as image classification \cite{krizhevsky2012imagenet} and language processing \cite{mikolov2013distributed} has prompted research towards multimodal tasks \cite{antol2015vqa, dhall2015video}  and virtual environments \cite{wu2018building, brodeur2017home, kolve2017ai2}. This is substantiated by embodiment theories in cognitive science that have argued for agent learning to be interactive and multimodal, mimicking key aspects of human learning \cite{landau1998object, mathematical_kids2012}. To foster and measure progress in such virtual environments, new tasks have been introduced, one of them being Embodied Question Answering (EmbodiedQA) \cite{das2018embodied}. \\ 


The EmbodiedQA task requires an agent to intelligently navigate in a simulated household environment \cite{wu2018building} and answer questions through egocentric vision. Concretely, an agent is spawned at a random location in an environment (a house or building) and asked a question (e.g. ‘What color
is the car?’). The agent perceives its environment through
first-person egocentric vision and can perform a few atomic
actions (move-forward, turn, strafe, etc.). The goal of the
agent is to intelligently navigate the environment and gather visual information necessary for answering the question.  
Subsequent to the introduction of the task, several methods have been introduced to solve the EmbodiedQA task \cite{das2018embodied, dasneural}, using some combination of reinforcement learning, behavior cloning and hierarchical control. Apart from using the question and images from the environment, these methods also rely on varying degrees of expert supervision such as shortest path demonstrations and subgoal policy sketches.

In this work, we evaluate simple question-only baselines that never see the environment and receive no form of expert supervision. We examine whether existing methods outperform baselines designed to solely capture dataset bias, in order to better understand the performance of these existing methods.
To our surprise, blindfold baselines achieve state-of-the-art performance on the EmbodiedQA task, except in the case when the agent is spawned extremely close to the object. Even in the latter case, blindfold baselines perform surprisingly close to existing state-of-the-art methods. We note that this finding is reminiscent of several recent works in both Computer Vision and Natural Language Processing, where researchers have found that statistical irregularities in the dataset can enable degenerate methods to perform surprisingly well \cite{poliak2018hypothesis, jabri2016revisiting, gururangan2018annotation, kaushik2018much}. \\

Our findings suggest that current EmbodiedQA models are ineffective at leveraging the context from the environment, in fact this context or embodiment in the environment can negatively hamper them. We hope comparison with our baseline results can more effectively demonstrate how well a method is able to leverage embodiment in the environment. Upon further error analysis of our models and qualitative inspection of the dataset, we find that there exist biases in the EQAv1 dataset that allow blindfold models to perform so well. We acknowledge the active effort of \citet{das2018embodied} in removing some biases via entropy-pruning but note that further efforts might be necessary to fully correct these biases. 

\section{Related Work}
\textbf{EmbodiedQA Methods}: \citet{das2018embodied} introduced the PACMAN-RL+Q model which is bootstrapped with expert shortest-path demonstrations and later fine-tuned with REINFORCE \cite{williams1992simple}. This model consists of a hierarchical navigation module: a planner and a controller, and a question answering module that acts when the navigation module has given up control. In a later work, \citet{dasneural} introduce Neural Modular Control (NMC) which is a hierarchical policy network that operates over expert sub-policy sketches. The master and sub-policies are initialized with Behavior Cloning (BC), and later fine-tuned with Asynchronous Advantage Actor-Critic (A3C) \cite{mnih2016asynchronous}. 

\textbf{Dataset Biases and Trivial Baselines}: Many recent studies in language and vision show how biases in a dataset allow models to perform well on a task without leveraging the meaning of the text or image in the underlying dataset. A simple CNN-BoW model was shown to achieve state-of-the-art results \cite{jabri2016revisiting} on the Visual7W \cite{zhu2016visual7w} task while also performing surprisingly well compared to the most complex systems proposed for the VQA dataset \cite{antol2015vqa} and other joint vision and language tasks \cite{frome2013devise,belilovsky2017joint}.  Simple nearest neighbor approaches have been shown to perform well on image captioning datasets \cite{devlin2015exploring}. This phenomenon has also been observed in language processing tasks. On the Story-cloze task which was presented to evaluate common-sense reasoning, \citet{schwartz2017story} achieved state-of-the-art performance by ignoring the narrative and training a linear classifier with features related to the writing style of the two potential endings, rather than their content. Similar observations were found on the Natural Language Inference (NLI) datasets, where methods ignoring the context and relying only on the hypothesis perform remarkably well \cite{poliak2018hypothesis, gururangan2018annotation}. Most recently, question-only and passage-only baselines on several QA datasets highlighted similar issues \cite{kaushik2018much}.

\section{Methods}

\vspace{-.75cm}
\begin{figure}[h!]
  \begin{minipage}[b]{0.525\textwidth}
  \includegraphics[width=.9\textwidth,height=2.35cm]{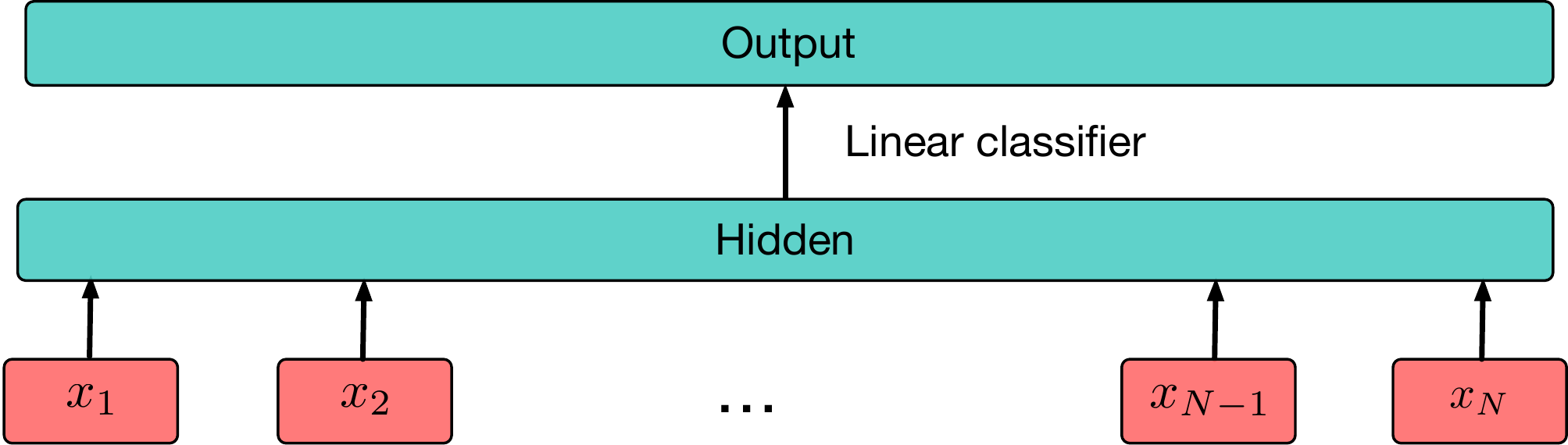}
  \vspace{.55cm}
  \caption{Model architecture for a
sentence with $N$ word vectors $x_1, \dots , x_N$ . The
embeddings are averaged to form the hidden variable. Figure adapted from \citet{joulin2016bag}.}
    \label{fig:1}
    \end{minipage}
    \hfill
   \begin{minipage}[b]{0.425\textwidth}
    \includegraphics[width=0.9\textwidth]{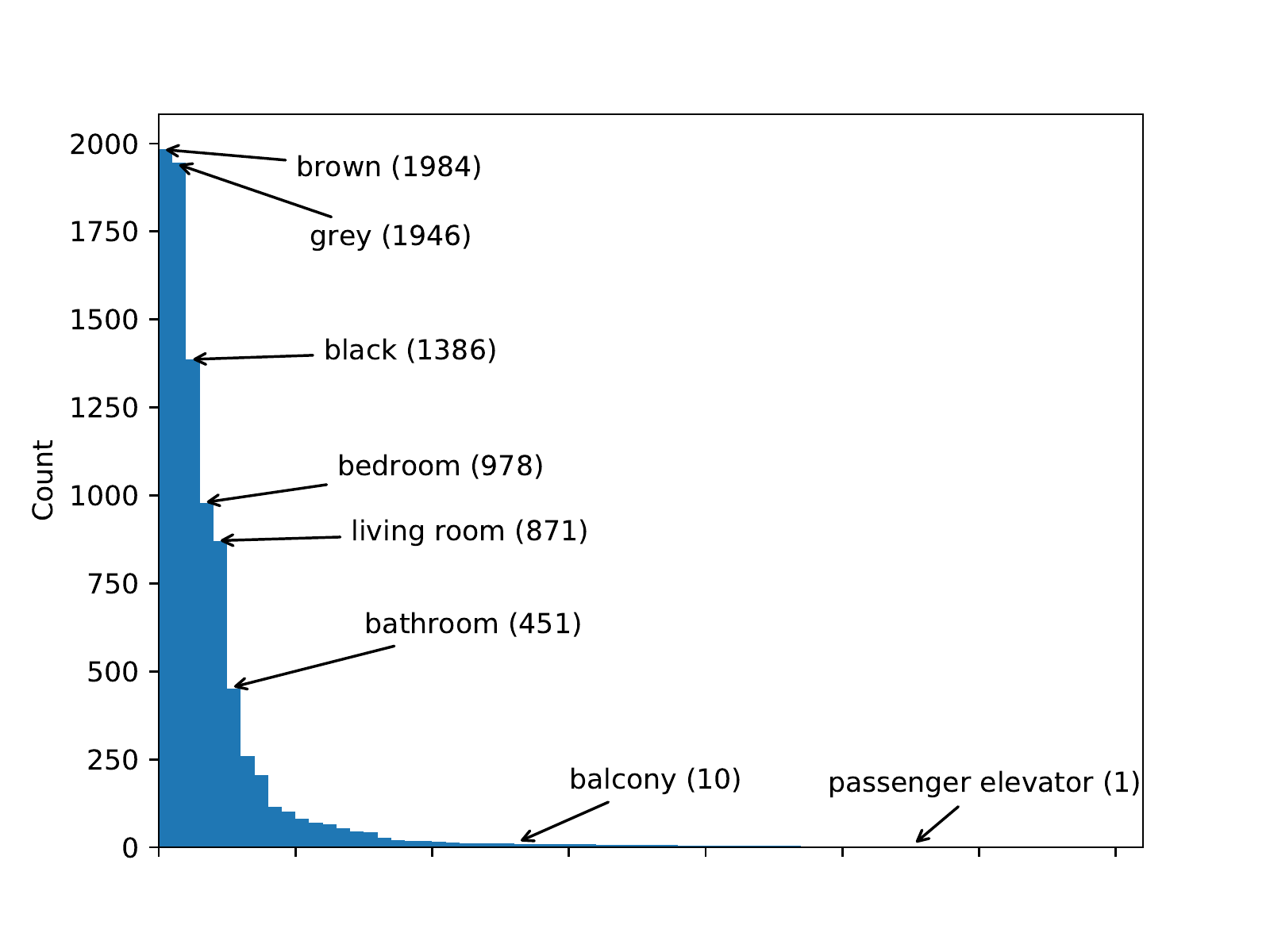}
  \caption{Frequency of each answer in the entire EQAv1 dataset. We observe that answers do not appear equally in the dataset, and are biased toward a select few.}
     \label{fig:2}
   \end{minipage}
\end{figure}

\paragraph{Average BOW Embedding}
We use a simple linear classifier as described in \cite{ren2015exploring, joulin2016bag, bengio2003neural}, which takes word level embeddings and averages them to construct the question representation.  We first perform a look-up over an embedding matrix for each word to get individual word representations. These word representations are then averaged into a text representation, which is in turn fed to a linear classifier. This architecture is similar to the \texttt{fastText} model of \cite{joulin2016bag}. It is also a common and strong baseline in language and vision and language tasks \cite{joulin2016bag,ren2015exploring}.  We use
the softmax function $f$ to compute the probability distribution over the predefined classes. The training criterion minimizes the negative log-likelihood over the classes.

\paragraph{Nearest Neighbor Answer Distribution (NN-AnswerDist)} This method attempts to answer purely based on the per-question answer distribution of the training set. For an input question we find either the identical question in the training set or if one doesn't exist the nearest matching question (based on number of shared words). We then select the most likely answer for the training set. Performance on this baseline is directly indicative of the bias in answer distributions in the dataset. We note that for EQAv1 almost all questions in the validation and test sets are present in the training set. 

\section{Experiments}
\textbf{EQAv1 Dataset} The EQAv1 dataset consists of 8985 questions split across 7190, 862 and 933 questions among training, validation and test sets respectively. The questions are generated via functional templates and are of following forms: 
\begin{description}
    \item \texttt{location}: \emph{'What room is the <OBJ> located in?'}
    \item \texttt{color}: \emph{'What color is the <OBJ>?'}
    \item \texttt{color\_room}: \emph{'What color is the <OBJ> in the <ROOM>?'}
    \item \texttt{preposition}: \emph{'What is <on/above/below/next-to> the
<OBJ> in the <ROOM>?'}
\end{description}
The answers span across 72 different categories of color, location and objects. We note that there are only 2 questions in the validation set, and 6 questions in the test set that are not in the training set. This limits the ability to test how well an agent generalizes across unseen combinations of rooms/objects/colors.  To get rid of peaky answers, an entropy pruning method was applied by \cite{das2018embodied} where questions with normalized entropy below 0.5 were excluded. However this still leaves an uneven answer distribution that can be exploited. 

\textbf{Training Details\footnote{Code for reproducing the experiments is available at https://github.com/ankeshanand/blindfold-baselines-eqa}} We evaluate the efficacy of our proposed baselines on the EQAv1 dataset. For the BoW model, we initialize the embeddings with Glove vectors \cite{pennington2014glove} of size 100, which are allowed to be fine-tuned during the training procedure. We use the Adam optimizer (batch-size of 64) with a learning rate of $5e^{-3}$ which is annealed via a scheduling mechanism based on plateaus in the validation loss. The training procedure is run for $200$ epochs and we use the checkpoint with minimum validation loss to compute accuracy on the test set. The NN-AnswerDist and the Majority baselines are self-descriptive and there are no specific training details that we apply. We also train the \cite{das2018embodied} text embedding model (an LSTM) with the optimization settings described in \cite{das2018embodied} for 200 epochs.

\textbf{Results} Detailed results are reported in Table~\ref{tab:results}. Following \citet{dasneural}, we report the agent's top-1 accuracy on the test set when spawned 10, 20 and 50 steps away from the goal, denoted as \emph{$T_{10}$} , \emph{$T_{20}$} and \emph{$T_{50}$} respectively. Since the performance of blindfold baselines are not affected based on where the agent is spawned, their accuracy is same across \emph{$T_{10}$} , \emph{$T_{20}$} and \emph{$T_{50}$}. We observe that the BoW model outperforms all existing methods except NMC(BC+A3C) in the case where agent is spawned very close to the target. The Nearest Neighbour method also does pretty well, and only falls behind to PACMAN (BC+REINFORCE) and NMC(BC+A3C) in the \emph{$T_{10}$} case. The difference in performance b/w the Nearest Neighbour method and BoW is primarily due to the fact that the BoW method leverages validation metrics more effectively, uses distributed word representations and differs in optimization. We also observe that the majority baseline achieves an accuracy of only $17.15\%$, suggesting that the other question-only baselines leverage dataset biases separate from class modes. For completeness, we also include a question only baseline derived directly from the EmbodiedQA codebase, which uses only the Question LSTM in the PACMAN model, termed as PACMAN Q-only (LSTM). Note that we only compare the top-1 accuracy of different methods here, and not the navigation performance since it's not directly applicable to these blindfold baselines.

To better understand the exact bias exploited by the text only models we observe that (a) The questions from training set are largely repeated in the validation and test set, with only 2 and 6 questions being unique to them respectively. As noted earlier, this means that models don't need to generalize across unseen combinations of rooms/objects/colors to perform well on this task (b) Despite entropy-pruning, there is a noticeable bias in the answer distribution of EQAv1 questions (see \cite[Appendix A]{das2018embodied}). Our results on the Nearest Neighbour baseline confirm this source of bias and explain largely the text model performance. 

Viewing these results holistically, we conclude that current methods for the EmbodiedQA task are not effective at using context from the environment, and in fact this negatively hampers them. This shows that there is room for building new models that leverage the context and embodiment in the environment.

\textbf{Oracles:} We now examine whether the EQAv1 dataset and the proposed oracle navigation can improve over pure text baselines, to leverage visual information in the most ideal case. We reproduce the settings for training the VQA model\footnote{We use the software provided by the authors https://github.com/facebookresearch/EmbodiedQA}. Specifically we train the VQA model described in \cite{dasneural} on the last 5 frames of oracle navigation for 50 epochs with ADAM and a learning rate of $3e-4$ using batch size 20. We observe the accuracy is improved over text baselines in this unrealistic setting, but the use of this model with navigation in PACMAN reduces performance to below the text baselines. For completeness we benchmark an oracle with our BoW embedding model in place of the LSTM with all other settings kept constant. As noted in \cite{das2018embodied}, we re-iterate that these oracles are far from perfect, as they may not contain
the best vantage or context to answer the question. 


\begin{table}[h!]
\label{tab:results}
\centering
\begin{tabular}{c|c|c|c|c}
           & $T_{10}$ & $T_{20}$ & $T_{50}$ & $T_{any}$ \\\hline
        \textbf{Navigation + VQA}  & & &\\\hline
     PACMAN (BC) \cite{das2018embodied}&  $48.48$ & $40.59$  & $39.87$ & N/A \\\hline
     PACMAN (BC+REINFORCE)\cite{das2018embodied}&  $50.21$ & $42.26$  & $40.76$ & N/A\\\hline
     NMC (BC) \cite{dasneural}&  $43.14$ & $41.96$  & $38.74$ & N/A \\\hline
     NMC (BC+A3C) \cite{dasneural}&  $\mathbf{53.58}$ & $46.21$  & $44.32$ & N/A
     \\ \hline \midrule
          \textbf{Question only}  & & &\\\hline
          Majority&  $17.15$ & $17.15$  & $17.15$ & $17.15$ \\\hline
    Nearest Neighbor Answer&  $48.45$ & $48.45$  & $48.45$ & $48.45$ \\\hline
     BOW&  $50.34$  & $\mathbf{50.34}$ & $\mathbf{50.34}$ & $\mathbf{50.34}$ \\\hline
     
     PACMAN Q-only (LSTM) (*)&  $46.07$ &  $46.07$  & 
     $46.07$&  $46.07$ \\\hline\midrule
\end{tabular} \\

\begin{tabular}{c|c}
     \textbf{Oracle VQA system}  & \\\hline
     PACMAN VQA-Only \cite{das2018embodied} (*)&  $55.9$  \\\hline
     BOW-CNN VQA-Only &  $56.5$  \\\hline
\end{tabular}
\caption{ We compare to the published results from \cite{dasneural} for agent spawned at various steps away from the target: 10, 30, 50, and anywhere in the environment. Question-only baselines outperform Navigation+VQA methods except when spawned 10 steps from the target object. A VQA-only system with oracle navigation can improve on a pure text baseline but isn't effective when combined with navigation. (*) indicates our reproduction of the model described in \cite{das2018embodied}
}
\end{table}

\textbf{Error Analysis}: To better understand the shortcomings and limitations, we perform an error analysis of the one of the runs of the BoW model on different question types: 
\begin{table}[h!]
\centering
\begin{tabular}{ c | c | c }
Preposition & Location & Color \\ \hline
 9.09 & 51.72 & 53.31 \\ \hline
\end{tabular}
\caption{Accuracy of the BoW model on different question types}
\end{table}
\vspace{-0.5cm}
Here, the color category subsumes \texttt{color} and 
\texttt{color\_room} both. The particularly low accuracy on preposition questions is due to the fact that there exist very few questions of this type in the training set ($2.44\%$), and the entropy of answer distribution in this class is much higher compared to color and location question types.

\section{Conclusion}
We show that simple question only baselines largely outperform or closely compete with existing methods on the EmbodiedQA task.
Our results indicate existing models are not able to convincingly use sensory inputs from the environment to perform question answering,
although they have been demonstrated some ability navigate toward the object of interest. Besides providing a benchmark score for future researchers working on this task, our results suggest considerations for future dataset and task construction in EQA and related tasks.

\section*{Acknowledgements}
We are grateful for the collaborative research environment provided by Mila. We also thank CIFAR for research funding and computing support. We further thank NVIDIA for donating a DGX-1 and Tesla
K80 used in this work. Lastly, we thank Sandeep, Rithesh, Abhishek and Catalina for reviewing earlier drafts of this work.

\bibliography{references}
\bibliographystyle{abbrvnat}

\small
\end{document}